\documentclass[15pt]{article}

\usepackage{arxiv}
\usepackage[utf8]{inputenc} 
\usepackage[hidelinks]{hyperref}
\usepackage{booktabs}       
\usepackage{amsfonts}       
\usepackage{nicefrac}       
\usepackage{microtype}      
\usepackage{graphicx}
\usepackage[numbers]{natbib}
\usepackage{doi}
\usepackage{relsize}
\usepackage{amssymb}
\usepackage{amsmath}
\usepackage{lineno}
\usepackage{multirow}
\usepackage{bigints}
\usepackage{amsbsy}
\usepackage{dcolumn}
\usepackage{bm}
\usepackage{cleveref}       
\usepackage[utf8]{inputenc}
\usepackage[T1]{fontenc}
\usepackage{mathptmx}
\usepackage{etoolbox}
\usepackage{xcolor}
\usepackage{subfig}
\usepackage{calrsfs}
\usepackage{bm}
\usepackage{booktabs}
\usepackage{multirow}
\usepackage{siunitx}
\DeclareMathAlphabet{\pazocal}{OMS}{zplm}{m}{n}

\def \PC {\textrm{\scalebox{1.05}{PC}$^2$}}

\usepackage[utf8]{inputenc}

\title{Polynomial chaos expansion for operator learning}

\date{}

\newif\ifuniqueAffiliation
\uniqueAffiliationtrue

\ifuniqueAffiliation 
\author{
        Himanshu Sharma\\
	Department of Civil and Systems Engineering\\
	  Johns Hopkins University\\
	Baltimore, USA  \\
	\And
	Luk{\'a}{\v s} Nov{\'a}k\\
	Department of Civil Engineering\\
	Brno University of Technology\\
        Brno, Czech Republic\\
        \And
        Michael Shields \thanks{Corresponding author;\  email: \texttt{michael.shields@jhu.edu}} \\
	Department of Civil and Systems Engineering\\
	  Johns Hopkins University\\
	Baltimore, USA  \\
 }

\begin{document}
\maketitle
\begin{abstract}

Operator learning (OL) has emerged as a powerful tool in scientific machine learning (SciML) for approximating mappings between infinite-dimensional functional spaces. One of its main applications is learning the solution operator of partial differential equations (PDEs). While much of the progress in this area has been driven by deep neural network-based approaches such as Deep Operator Networks (DeepONet) and Fourier Neural Operator (FNO), recent work has begun to explore traditional machine learning methods for OL. In this work, we introduce polynomial chaos expansion (PCE) as an OL method. PCE has been widely used for uncertainty quantification (UQ) and has recently gained attention in the context of SciML. For OL, we establish a mathematical framework that enables PCE to approximate operators in both purely data-driven and physics-informed settings. The proposed framework reduces the task of learning the operator to solving a system of equations for the PCE coefficients. Moreover, the framework provides UQ by simply post-processing the PCE coefficients, without any additional computational cost. We apply the proposed method to a diverse set of PDE problems to demonstrate its capabilities. Numerical results demonstrate the strong performance of the proposed method in both OL and UQ tasks, achieving excellent numerical accuracy and computational efficiency.

\end{abstract}


\section{Introduction}

Operator learning (OL) has emerged as an influential field within scientific machine learning (SciML), focusing on learning mappings between infinite-dimensional functional spaces. In particular, OL has demonstrated significant potential in approximating the solution operator of partial differential equations (PDEs) by effectively mapping initial conditions, boundary conditions, source terms, and/or parameters to the corresponding PDE solution. The majority of existing work in OL has focused on neural network (NN) based approaches, with frameworks such as Deep Operator Networks (DeepONet) \cite{lu2021learning}, Fourier Neural Operators (FNO) \cite{li2020fourier}, Wavelet Neural Operators (WNO) \cite{tripura2023wavelet}, Laplace Neural Operators (LNO) \cite{cao2024laplace}, Convolutional Neural Operators (CNO) \cite{raonic2023convolutional}, and Graph Neural Operators (GNO) \cite{li2020neural}. These models and their many variants have demonstrated significant potential across a wide range of scientific applications. 

Recently, there has been growing research interest in exploring traditional machine learning (ML) methods within the OL framework. Particularly, Gaussian Processes (GPs) and other kernel-based approaches have demonstrated strong performance, especially in data-scarce settings where NN-based approaches may struggle. These methods offer interpretable solutions with built-in uncertainty quantification, as well as theoretical and computational guarantees \cite{batlle2024kernel, mora2025operator, nelsen2024operator}. Furthermore, hybrid approaches that integrate traditional ML methods into neural operator networks have also shown improved performance. For example, Lowery et al. \cite{lowery2024kernel} proposed the Kernel Neural Operator (KNO), which incorporates trainable kernel functions into neural operator architectures and achieves competitive accuracy with fewer parameters compared to FNO and GNO. Similarly, a hybrid GP/NN framework \cite{mora2025operator}, which combines the expressiveness of NNs with the uncertainty quantification and interpretability of GPs, has shown enhanced performance. These traditional and hybrid frameworks offer a promising direction for OL, particularly in terms of computational efficiency, convergence guarantees, reduced data and resource requirements, and uncertainty quantification.

In this work, we introduce polynomial chaos expansion (PCE) for OL. PCE is a widely used method in the field of uncertainty quantification (UQ), where it approximates the solution of a stochastic system using a spectral expansion of orthogonal multivariate polynomials with deterministic coefficients. The choice of orthogonal polynomials depends on the probability distributions of the input stochastic variables, typically following the Wiener–Askey scheme \cite{xiu2002wiener}. Alternatively, for arbitrary input distributions, the polynomials can be constructed using approaches such as those proposed in \cite{soize2004physical}. Recently, Bahmani et al. \cite{bahmani2025neural} proposed a neural chaos method, where the orthogonal polynomial basis functions are replaced with NN parameterized basis functions in a purely data-driven manner. Once the basis is selected, the deterministic coefficients in PCE can be computed using either intrusive approaches, such as the stochastic Galerkin (SG) method \cite{ghanem2003stochastic}, or non-intrusive approaches, such as the stochastic collocation method \cite{xiu2016stochastic}. Non-intrusive methods are commonly categorized into three main types: interpolation \cite{xiu2005high,narayan2012stochastic}, regression \cite{berveiller2006stochastic}, and pseudo-projection methods \cite{xiu2007efficient}. The resulting PCE coefficients can then be post-processed to efficiently estimate various response statistics, such as moments and sensitivities with respect to the input stochastic variables \cite{sudret2008global, novak2022distribution,NOVAK2025110594}.

PCE has also been recently employed as an ML method in a purely data-driven setting, demonstrating competitive performance compared to NNs \cite{torre2019data}. In SciML, we have recently proposed physics-constrained polynomial chaos expansion (\PC{}) \cite{novak2024physics, sharma2024physics} -- which incorporates physical constraints into the PCE framework by solving a constrained least squares regression problem to determine the PCE coefficients -- and manifold PCE (m-PCE)~\cite{kontolati2022manifold,kontolati2022survey}, which develops data-driven PCE on low-dimensional manifolds for high-dimensional UQ and has demonstrated competitive performance to DeepONet for complex PDEs~\cite{kontolati2023influence}. While \PC{} and m-PCE are not explicitly presented as OL methods, they lay the groundwork for the current work, which formally establishes PCE as an OL method. 

This work presents the mathematical formulation for approximating the solution operator of a given PDE using PCE. The proposed framework reduces the task of OL to solving a system of equations for the PCE coefficients. Furthermore, the proposed method is capable of learning operators in both purely data-driven and physics-informed settings. Our framework is mathematically rigorous, easy to implement, and computationally efficient. Moreover, it provides UQ at no additional computational cost. We demonstrate the effectiveness of the proposed method in both OL and UQ tasks by applying it to a diverse set of PDE problems.

\section{Methodology}
\label{sec:methodology}

Consider a general operator learning task, where the goal is to learn an operator $\pazocal{G}$ mapping input functions $u$ defined over a domain $\pazocal{D}$ to output functions $s$ defined over a domain $\pazocal{R}$:
\begin{equation}
    \pazocal{G}: \pazocal{U} \rightarrow \pazocal{S}, \quad u(x) \mapsto s(y),
\end{equation}
where $\pazocal{U}$ and $\pazocal{S}$ are function spaces (e.g., Banach or Hilbert spaces), $x \in \pazocal{D} \subseteq \mathbb{R}^{d_x}$ and $y \in \pazocal{R} \subseteq \mathbb{R}^{d_y}$. Here, $d_x$ and $d_y$ denote the dimensionality of the input and output domains, respectively.

The learning goal is to find an approximate operator $\pazocal{G}_\theta$, parameterized by $\theta$, that minimizes the discrepancy between predicted and true outputs over the given dataset $\left\{\left(u^{(i)}, s^{(i)}\right)\right\}_{i=1}^N, \text { with } s^{(i)}=\pazocal{G}\left(u^{(i)}\right)$:
\begin{equation}\label{eq:OL-optimization}
\theta^*=\arg \min _\theta \frac{1}{N} \sum_{i=1}^N \pazocal{L}\left(s^{(i)}, \pazocal{G}_\theta\left(u^{(i)}\right)\right),
\end{equation}
where $\pazocal{L}$ is a suitable loss function (e.g., mean squared error) or any other metric suitable for the operator-learning task at hand.

In scientific and engineering applications, a common scenario involves learning an operator that is implicitly defined by a physical system governed by partial differential equations (PDEs).
Consider a general PDE defined on a spatial domain $\pazocal{D} \subseteq \mathbb{R}^d$ and time domain $t \in$ $[0, T]$ as
\begin{equation}\label{eq:general-pde}
\begin{aligned}
&\pazocal{N}_{\nu}[s(\mathbf{x}, t)]=f(\mathbf{x}, t), \quad \mathbf{x} \in \pazocal{D},\ t \in[0, T],\\
&\pazocal{B}[s(\mathbf{x}, t)]=h(\mathbf{x}, t), \quad \mathbf{x} \in \partial \pazocal{D},\ t \in[0, T],\\
& s(\mathbf{x}, 0)=g(\mathbf{x}), \quad \mathbf{x} \in \pazocal{D},
\end{aligned}
\end{equation}
where $s(\mathbf{x}, t)$ is the unknown PDE solution, $\pazocal{N}_\nu$ represents a general differential operator parameterized by $\nu$, $\pazocal{B}$ corresponds to the boundary operator, and $\partial \pazocal{D}$ denotes the boundary of the spatial domain. The functions $f(\mathbf{x}, t),\ h(\mathbf{x}, t)$, and $g(\mathbf{x})$ represent the source term, boundary condition, and initial condition, respectively. In the context of operator learning, any combination of these inputs, i.e., $f$, $h$, $g$, and $\nu$, may collectively form the input function $u$. The goal is to learn the mapping from $u$ to the corresponding PDE solution $s$.  

For the task of learning the operator, these input functions are usually sampled from random fields $u(\mathbf{x},t,\omega)$, $\omega \in \Omega$, and $\Omega$ denotes the underlying sample space of the probability space ($\Omega, \mathcal{F}, \mathcal{P})$ where $\mathcal{F}$ is the sigma-algebra of events and $\mathcal{P}$ is the associated probability measure. Once the operator is learned, it can provide the PDE solution corresponding to any given realization of the input random fields. This problem setup naturally fits within the framework of PCE, where PCE is used to learn the operator.

We can express the solution operator using a PCE as
\begin{equation}\label{eq:PCE-OL}
\pazocal{G}(u(\mathbf{x},t,\boldsymbol{\xi}\left(\omega\right)))(\mathbf{x},t)  \approx \hat{s}(\mathbf{x},t,\boldsymbol{\xi}\left(\omega\right)) = \sum_{\boldsymbol{\alpha} \in \pazocal{A}} s_{\boldsymbol{\alpha}}\left(\mathbf{x},t\right) \Psi_{\boldsymbol{\alpha}}\left(\boldsymbol{\xi}\left(\omega\right)\right),
\end{equation}
where $\boldsymbol{\xi} = \left(\xi_{1}, \xi_{2}, \ldots, \xi_{r}\right) \in \pazocal{D}_{\boldsymbol{\xi}}\subseteq \mathbb{R}^r$ is a vector of independent random variables with joint PDF $p_{\boldsymbol{\xi}}(\boldsymbol{\xi})=$ $\prod_{i=1}^r p_{\xi_i}\left(\xi_i\right)$. The multivariate polynomials $\left\{\Psi_\alpha(\boldsymbol{\xi})\right\}_{\alpha \in \pazocal{A}}$ are constructed by taking tensor-products of univariate orthogonal polynomials $\psi_{\alpha_i}\left(\xi_i\right)$. They satisfy the orthogonality relation
$$
\int_{\pazocal{D}_{\boldsymbol{\xi}}} \Psi_\alpha(\boldsymbol{\xi}) \Psi_\beta(\boldsymbol{\xi}) p_{\xi}(\boldsymbol{\xi}) d \boldsymbol{\xi}=0 \quad(\boldsymbol{\alpha} \neq \boldsymbol{\beta}),
$$
and thus form an orthogonal basis of the weighted Hilbert space $L^2\left(\pazocal{D}_{\boldsymbol{\xi}}, p_{\boldsymbol{\xi}}(\boldsymbol{\xi}) d \boldsymbol{\xi}\right)$. These basis functions are typically selected according to the Wiener-Askey scheme \cite{xiu2002wiener}, but can also be constructed for arbitrary distributions of $\boldsymbol{\xi}$ \cite{soize2004physical}. The coefficients $s_{\boldsymbol{\alpha}}\left(\mathbf{x},t\right)$ are deterministic functions of the spatio-temporal variables and must be determined. The index set $\pazocal{A}$ is a finite set of multi-indices with cardinality $P$. 

A standard truncation scheme includes all multivariate polynomials in the
$r$ input variables whose total degree does not exceed $p$, forming the index set:
\begin{equation}
    \pazocal A=\left\{\boldsymbol{\alpha} \in \mathbb{N}^r:{\|\boldsymbol{\alpha}\|}_{1} \leq p\right\} .
\end{equation}

The cardinality of the truncated index set $ \pazocal A $ is given by
\begin{equation}\label{eq.:Cardinality PCE}
 \mathrm{card} \: \pazocal A= \frac{\left( r+p \right)!}{r! \: p!}\equiv P . 
\end{equation}
Other truncation schemes can be employed to reduce the cardinality of the index set, for example, by limiting the number of interaction terms through approaches such as hyperbolic truncation \cite{blatman2011adaptive}.

Next, we represent the deterministic coefficient function, $s_{\boldsymbol{\alpha}}\left(\mathbf{x},t\right)$, as a spectral expansion of orthogonal polynomials given as
\begin{equation}\label{eq.:ortho-polynomial-exp}
    s_{\boldsymbol{\alpha}}\left(\mathbf{x},t\right) = \sum_{\boldsymbol{\beta} \in \pazocal{B}} c_{\boldsymbol{\alpha}\boldsymbol{\beta}} \Phi_{\boldsymbol{\beta}}\left(\mathbf{x},t\right),
\end{equation}
where $\Phi_{\boldsymbol{\beta}}\left(\mathbf{x},t\right)$ are multivariate orthogonal polynomial basis functions, constructed as tensor products of one-dimensional orthogonal polynomials with respect to some weight function, $w(x)$, defined over a specified interval for the spatial and temporal dimensions (e.g, Legendre polynomials defined on $[-1,1]$ with $w(x)=1$), and $c_{\boldsymbol{\alpha}\boldsymbol{\beta}}$ are the corresponding scalar coefficients. 

We adopt the same truncation scheme as used for Eq.~\eqref{eq:PCE-OL}, i.e.,  
$
\pazocal B^{d+1,q}=\left\{\boldsymbol{\beta} \in \mathbb{N}^{d+1}:{\|\boldsymbol{\beta}\|}_{1} \leq q\right\},
$ 
where $q$ denotes the total degree of the expansion, $d$ is the number of spatial dimensions, which can be at most three, along with one temporal dimension. The cardinality of set $\pazocal{B}$ is given by
\begin{equation}
    Q= \frac{\left( d+1+q \right)!}{(d+1)! \: q!}.
\end{equation}

Substituting Eq.~\eqref{eq.:ortho-polynomial-exp} in Eq.~\eqref{eq:PCE-OL}, we get the expansion of the solution operator as
\begin{equation}\label{eq:PCE}
     \hat{s}(\mathbf{x},t,\boldsymbol{\xi}) = \sum_{\boldsymbol{\alpha} \in \pazocal{A}} \sum_{\boldsymbol{\beta} \in \pazocal{B}} c_{\boldsymbol{\alpha}\boldsymbol{\beta}} \Phi_{\boldsymbol{\beta}}\left(\mathbf{x},t\right) \Psi_{\boldsymbol{\alpha}}(\boldsymbol{\xi)}.
\end{equation}

Rewriting Eq.~\eqref{eq:PCE} in matrix form, we get
\begin{equation}\label{eq:PCE-matrix form}
    \hat{\mathbf{S}}=\boldsymbol{\Phi} \mathbf{C} \boldsymbol{\Psi},
\end{equation}
where
\begin{equation}
\boldsymbol{\Phi}=\left[\begin{array}{ccc}
\Phi_1\left((\mathbf{x},t)^{(1)}\right) & \ldots & \Phi_Q\left((\mathbf{x},t)^{(1)}\right) \\
\vdots & \ddots & \vdots \\
\Phi_1\left((\mathbf{x},t)^{(n)}\right) & \ldots & \Phi_Q\left((\mathbf{x},t)^{(n)}\right)
\end{array}\right],
\end{equation}

\begin{equation}
\boldsymbol{\Psi}=\left[\begin{array}{ccc}
\Psi_{1}\left(\boldsymbol{\xi}^{(1)}\right) & \ldots & \Psi_{1}\left(\boldsymbol{\xi}^{(N)}\right) \\
\vdots & \ddots & \vdots \\
\Psi_{P}\left(\boldsymbol{\xi}^{(1)}\right) & \ldots & \Psi_{P}\left(\boldsymbol{\xi}^{(N)}\right)
\end{array}\right].
\end{equation}
Here, $\boldsymbol{\Phi}\in \mathbb{R}^{n \times Q}$, $\boldsymbol{\Psi}\in \mathbb{R}^{P \times N}$, $n$ denotes the number of spatio-temporal coordinates at which the operator is evaluated, and $N$ is the number of samples of the input functions.
The unknowns are the coefficients $\mathbf{C} \in \mathbb{R}^{Q \times P}$, which represent $\theta$ in Eq.~\eqref{eq:OL-optimization}. 

The OL task reduces to learning the coefficient matrix of the PCE. Given the labeled training data, the coefficient can be obtained by minimizing the residual as
\begin{equation} \label{eq:PCE-optimization}
\mathbf{C}^{*} = \dfrac{1}{n\cdot N} 
 \underset{\tilde{\mathbf{C}}}{\LARGE{\arg \min }} \Big\| \mathbf{S}- \hat{\mathbf{S}} \Big\|^2,
\end{equation}
where $\mathbf{S}$ is the true solution matrix and $\hat{\mathbf{S}}$ is the approximate PCE-based solution matrix, as defined in Eq.~\eqref{eq:PCE-matrix form}.

The closed-form solution for Eq.~\eqref{eq:PCE-optimization} reads as
\begin{equation}\label{eq:PCE-optimal-coefficient}
    \mathbf{C}^{*} = \left(\boldsymbol{\Phi}^{\top}\boldsymbol{\Phi}\right)^{-1}\left(\boldsymbol{\Phi}^{\top}\mathbf{S}\boldsymbol{\Psi}^{\top} \right)\left(\boldsymbol{\Psi}\boldsymbol{\Psi}^{\top}\right)^{-1}.
\end{equation}

For a unique solution to Eq.~\eqref{eq:PCE-optimal-coefficient}, it is required that $n\geq Q$ and $N\geq P$. Once $\mathbf{C}^{*}$ is obtained, substituting it back into Eq.~\eqref{eq:PCE-matrix form} yields the approximate solution for a given test input function at any desired spatial-temporal coordinates.

Next, we learn the operator in a physics-informed manner, i.e., without requiring labeled training data. We refer to this approach as the physics-constrained polynomial chaos expansion (\PC{}). It is worth noting that the present \PC{} formulation differs from our previous work on this topic \cite{novak2024physics, sharma2024physics}, and is significantly more efficient and better tailored for operator learning. In this work, the stochastic and spatio-temporal basis functions are separated, leading to a far more compact expansion and efficient solution. 

Given the general PDE described in Eq.~\eqref{eq:general-pde}, we replace the unknown PDE solution with the PCE approximation from Eq.~\eqref{eq:PCE}. This gives 
\begin{equation}\label{eq:PC^2-PDE}
\begin{aligned}
    &\pazocal{N}_{\nu}[\hat{s}(\mathbf{x}, t, \boldsymbol{\xi})] = \sum_{\boldsymbol{\alpha} \in \pazocal{A}} \sum_{\boldsymbol{\beta} \in \pazocal{B}} c_{\boldsymbol{\alpha}\boldsymbol{\beta}} \pazocal{N}_{\nu}\left[\Phi_{\boldsymbol{\beta}}\left(\mathbf{x},t\right)\right] \Psi_{\boldsymbol{\alpha}}(\boldsymbol{\xi)} \approx f(\mathbf{x}, t, \boldsymbol{\xi}),\\
    &\pazocal{B}[\hat{s}(\mathbf{x}, t, \boldsymbol{\xi})] = \sum_{\boldsymbol{\alpha} \in \pazocal{A}} \sum_{\boldsymbol{\beta} \in \pazocal{B}} c_{\boldsymbol{\alpha}\boldsymbol{\beta}} \pazocal{B}\left[\Phi_{\boldsymbol{\beta}}\left(\mathbf{x},t\right)\right] \Psi_{\boldsymbol{\alpha}}(\boldsymbol{\xi)} \approx h(\mathbf{x}, t, \boldsymbol{\xi}),\\
    & \hat{s}(\mathbf{x}, 0, \boldsymbol{\xi}) = \sum_{\boldsymbol{\alpha} \in \pazocal{A}} \sum_{\boldsymbol{\beta} \in \pazocal{B}} c_{\boldsymbol{\alpha}\boldsymbol{\beta}} \Phi_{\boldsymbol{\beta}}\left(\mathbf{x}, 0\right) \Psi_{\boldsymbol{\alpha}}(\boldsymbol{\xi)} \approx g(\mathbf{x}, \boldsymbol{\xi}).
\end{aligned}
\end{equation}

The differential operators apply only to the physical basis functions, $\Phi_{\boldsymbol{\beta}}\left(\mathbf{x},t\right)$, as the constraints are defined with respect to the spatio-temporal variables rather than the stochastic variables. The constraints are enforced at randomly selected spatio-temporal virtual points for $N$ given realizations of the input random vector.
Let the number of virtual points corresponding to the PDE, BC and IC be $n_{\mathrm{PDE}}$, $n_{\mathrm{BC}}$ and $n_{\mathrm{IC}}$, respectively. From Eq.~\eqref{eq:PCE-matrix form} and Eq.~\eqref{eq:PC^2-PDE}, we can write the residuals as
\begin{equation}\label{eq:residuals}
\begin{aligned}
    \mathbf{R}_{\mathrm{PDE}} & = \boldsymbol{\Phi}_{\mathrm{PDE}} \mathbf{C} \boldsymbol{\Psi} - \mathbf{F}_{\mathrm{PDE}}, \\
    \mathbf{R}_{\mathrm{BC }} &= \boldsymbol{\Phi}_{\mathrm{BC}} \mathbf{C} \boldsymbol{\Psi} - \mathbf{F}_{\mathrm{BC}}, \\
    \mathbf{R}_{\mathrm{IC }} &= \boldsymbol{\Phi}_{\mathrm{IC}} \mathbf{C} \boldsymbol{\Psi} - \mathbf{F}_{\mathrm{IC}}, \\
\end{aligned}
\end{equation}
where matrices $\boldsymbol{\Phi}_{\mathrm{PDE}}\in \mathbb{R}^{n_{\mathrm{PDE}}\times Q}$, $\boldsymbol{\Phi}_{\mathrm{BC}}\in \mathbb{R}^{n_{\mathrm{BC}}\times Q}$ and $\boldsymbol{\Phi}_{\mathrm{IC}}\in \mathbb{R}^{n_{\mathrm{IC}}\times Q}$ are evaluated at virtual points corresponding to the PDE, BC and IC, respectively, after the application of the respective differential operators. The matrix $\boldsymbol{\Psi}\in \mathbb{R}^{P\times N}$ is evaluated for the given $N$ input realizations of the random vector. The matrices $\mathbf{F}_{\mathrm{PDE}}\in \mathbb{R}^{n_{\mathrm{PDE}}\times N}$, $\mathbf{F}_{\mathrm{BC}}\in \mathbb{R}^{n_{\mathrm{BC}}\times N}$ and $\mathbf{F}_{\mathrm{IC}}\in \mathbb{R}^{n_{\mathrm{IC}}\times N}$ contain the values of input functions corresponding to the PDE, BC, and IC, respectively, evaluated at their associated virtual points for each realization of the input random vector.

Now, using Eq.~\eqref{eq:residuals}, we can write the loss function for \PC{} as
\begin{equation}\label{eq:PC^2-loss function}
\begin{aligned}
L(\mathbf{C}) & =L_{\mathrm{PDE}}(\mathbf{C})+L_{\mathrm{BC}}(\mathbf{C})+L_{\mathrm{IC}}(\mathbf{C}) \\
& =\frac{1}{n_{\mathrm{PDE}}\cdot N}\left\|\mathbf{R}_{\mathrm{PDE}}\right\|^2+\frac{1}{n_{\mathrm{BC}}\cdot N}\left\|\mathbf{R}_{\mathrm{BC}}\right\|^2+\frac{1}{n_{\mathrm{IC}}\cdot N}\left\|\mathbf{R}_{\mathrm{IC}}\right\|^2.
\end{aligned}
\end{equation}

The coefficients can be learned by minimizing the loss function in Eq.~\eqref{eq:PC^2-loss function}, for which we set $\nabla L(\mathbf{C})=0$. For linear PDEs, this results in a system of linear equations, while for nonlinear PDEs, it leads to a system of nonlinear equations. Furthermore, a data loss term can be added to Eq.~\eqref{eq:PC^2-loss function} if training data is available. Thus, the task of operator learning reduces to solving a system of equations.

For linear PDEs, we get 
\begin{equation}
\begin{aligned}
    \nabla L(\mathbf{C}) =& \frac{2}{n_{\mathrm{PDE}}}\boldsymbol{\Phi}_{\mathrm{PDE}}^{\top}\left(\boldsymbol{\Phi}_{\mathrm{PDE}} \mathbf{C} \boldsymbol{\Psi}-\mathbf{F}_{\mathrm{PDE}}\right) \boldsymbol{\Psi}^{\top} + \frac{2}{n_{\mathrm{IC}}}\boldsymbol{\Phi}_{\mathrm{IC}}^{\top}\left(\boldsymbol{\Phi}_{\mathrm{IC}} \mathbf{C} \boldsymbol{\Psi}-\mathbf{F}_{\mathrm{IC}}\right) \boldsymbol{\Psi}^{\top} \\ +& \frac{2}{n_{\mathrm{BC}}} \boldsymbol{\Phi}_{\mathrm{BC}}^{\top}\left(\boldsymbol{\Phi}_{\mathrm{BC}} \mathbf{C} \boldsymbol{\Psi}-\mathbf{F}_{\mathrm{BC}}\right) \boldsymbol{\Psi}^{\top} = 0
\end{aligned}
\end{equation}

This gives 
\begin{equation}
\begin{aligned}
            &\left[\frac{2}{n_{\mathrm{PDE}}}\boldsymbol{\Phi}_{\mathrm{PDE}}^{\top}\boldsymbol{\Phi}_{\mathrm{PDE}} +  \frac{2}{n_{\mathrm{IC}}}\boldsymbol{\Phi}_{\mathrm{IC}}^{\top}\boldsymbol{\Phi}_{\mathrm{IC}} + \frac{2}{n_{\mathrm{BC}}} \boldsymbol{\Phi}_{\mathrm{BC}}^{\top}\boldsymbol{\Phi}_{\mathrm{BC}} \right] \mathbf{C} \boldsymbol{\Psi} \boldsymbol{\Psi}^{\top} \\
        &=\frac{2}{n_{\mathrm{PDE}}}\boldsymbol{\Phi}_{\mathrm{PDE}}^{\top}\mathbf{F}_{\mathrm{PDE}} \boldsymbol{\Psi}^{\top}+\frac{2}{n_{\mathrm{IC}}}\boldsymbol{\Phi}_{\mathrm{IC}}^{\top}\mathbf{F}_{\mathrm{IC}} \boldsymbol{\Psi}^{\top}+\frac{2}{n_{\mathrm{BC}}}\boldsymbol{\Phi}_{\mathrm{BC}}^{\top}\mathbf{F}_{\mathrm{BC}} \boldsymbol{\Psi}^{\top}
\end{aligned}
\end{equation}

The optimal coefficients are obtained as 
\begin{eqnarray}\label{eq:PC^2-optimal-coeff-linear}
    \begin{aligned}
  \mathbf{C^{*}} 
        &=\left[\frac{2}{n_{\mathrm{PDE}}}\boldsymbol{\Phi}_{\mathrm{PDE}}^{\top}\boldsymbol{\Phi}_{\mathrm{PDE}} + \frac{2}{n_{\mathrm{IC}}}\boldsymbol{\Phi}_{\mathrm{IC}}^{\top}\boldsymbol{\Phi}_{\mathrm{IC}} + \frac{2}{n_{\mathrm{BC}}} \boldsymbol{\Phi}_{\mathrm{BC}}^{\top}\boldsymbol{\Phi}_{\mathrm{BC}} \right]^{-1}\\
        &\left[\frac{2}{n_{\mathrm{PDE}}}\boldsymbol{\Phi}_{\mathrm{PDE}}^{\top}\mathbf{F}_{\mathrm{PDE}} \boldsymbol{\Psi}^{\top}+\frac{2}{n_{\mathrm{IC}}}\boldsymbol{\Phi}_{\mathrm{IC}}^{\top}\mathbf{F}_{\mathrm{IC}} \boldsymbol{\Psi}^{\top}+\frac{2}{n_{\mathrm{BC}}}\boldsymbol{\Phi}_{\mathrm{BC}}^{\top}\mathbf{F}_{\mathrm{BC}} \boldsymbol{\Psi}^{\top}\right]\left[\boldsymbol{\Psi} \boldsymbol{\Psi}^{\top} \right]^{-1}
    \end{aligned}
\end{eqnarray}

For nonlinear PDEs, we solve a nonlinear system of equations using Newton-Raphson as 
\begin{equation}\label{eq:PC^2-optimal-coeff-nonlinear}
    \mathbf{C}^{(k+1)}=\mathbf{C}^{(k)}-\left[\pazocal{H} L\left(\mathbf{C}^{(k)}\right)\right]^{-1} \nabla L\left(\mathbf{C}^{(k)}\right),
\end{equation}
where $\nabla L\left(\mathbf{C}^{(k)}\right) \in \mathbb{R}^{Q\times P}$ is the gradient of $L$ evaluated at the current iterate $\mathbf{C}^{(k)}$, and $\pazocal{H} L\left(\mathbf{C}^{(k)}\right) \in \mathbb{R}^{Q\times P}$ is the Hessian matrix of $L$ at $\mathbf{C}^{(k)}$. Closed-form expressions for the gradient and Hessian can be derived due to the polynomial structure of the expansion. We could also compute the optimal coefficients by minimizing the loss function using numerical optimization algorithms, such as gradient descent, ADAM \cite{kingma2014adam}, or L-BFGS.

Next, UQ can be efficiently performed by computing the first two moments at any desired spatio-temporal coordinate directly from the coefficient matrix $\mathbf{C}^*$, as follows:
\begin{equation}
\mu(\mathbf{x},t)=\mathbb{E}[\hat{\mathbf{S}}]=\boldsymbol{\Phi}(\mathbf{x},t) \mathbf{c}_0 \in\mathbb{R}^{n}
\end{equation}
where $\mathbf{c}_0$ is the first column of $\mathbf{C^{*}}$ matrix.

The covariance matrix is given as:
\begin{equation}
\boldsymbol{\Sigma}=\boldsymbol{\Phi}(\mathbf{x},t)\left(\mathbf{C^{*}} \operatorname{diag}(0,1, \ldots, 1) \mathbf{C^{*}}^{\top}\right) \boldsymbol{\Phi}^{\top}(\mathbf{x},t)\in\mathbb{R}^{n\times n}
\end{equation}

Additional output statistics, such as Sobol sensitivity indices \cite{ sudret2008global,sobol1990sensitivity}, can also be analytically derived from the coefficient matrix $\mathbf{C^{*}}$.

\section{Numerical results}

In this section, we demonstrate the capabilities of the PCE and \PC{} in OL and UQ by considering four diverse ODE and PDE problems. For all example problems, the input functions are sampled from a Gaussian random field (GRF), which is discretized using the truncated Karhunen–Loève (KL) expansion \cite{ghanem2003stochastic}. We generate the training and testing data using the FEM solutions corresponding to different realizations of the input function. The training data is used solely for training PCE and is not required for \PC{}. The performance of both PCE and \PC{} is evaluated on the same testing dataset. To validate the UQ results, we use Monte Carlo simulations (MCS) with 10,000 samples to compute reference estimates of the mean and standard deviation for all example problems. For OL, we assess the numerical accuracy of PCE and \PC{} by computing the mean square error (MSE) with respect to FEM solutions on the testing samples. For UQ, we evaluate performance using the mean absolute error (MAE) in the predicted mean and standard deviation over the given domain, compared to reference results from MCS.

Table \ref{tab:PCE-performance} summarizes the performance of PCE on all example problems for both OL and UQ. It can be observed that PCE excels in both OL and UQ tasks, providing exceptional numerical accuracy and computational efficiency. For OL, the MSE across all problems is on the order of $10^{-9}$ to $10^{-4}$, with training times of fractions of a second, which demonstrates excellent accuracy and efficiency of the PCE method. The UQ results show significantly lower MAE for both the mean and standard deviation across all examples, with a fraction of the computational cost compared to MCS, indicating superior performance. 

From these results, it is evident that PCE excels in both OL and UQ tasks. However, as discussed in Section \ref{sec:methodology}, the PCE solution (Eq.~\eqref{eq:PCE-optimal-coefficient}) is unique when $N\geq P$. Moreover, $P$ increases significantly with the PCE order ($p$) and the stochastic dimensionality ($r$) (Eq.~\eqref{eq.:Cardinality PCE}), necessitating a larger amount of training data, which may become computationally infeasible to obtain. For example, in the case of the 2D heat problem, we get $P = 2045$, and therefore we use $N=2500$ labeled training samples to ensure a unique solution. Acquiring this amount of labeled training data was computationally expensive ($\sim$7 h). Hence, it is essential to use \PC{}, which eliminates the need for labeled training data. 

Table \ref{tab:PC^2-performance} shows the \PC{} performance for both the OL and UQ tasks. As observed, the \PC{} achieves excellent numerical accuracy and computational efficiency for both OL and UQ across all example problems. \PC{} offers competitive performance relative to standard PCE without requiring any labeled training data, making it an ideal choice for OL and UQ, especially for high-dimensional problems. Thus, for all the example problems, we generate visualization plots using \PC{}. The details of the problem setup for each example are provided in the subsequent sections. 

The codes were written in Python using the packages UQpy \cite{tsapetis2023uqpy}, JAX \cite{bradbury2018jax}, and  FEniCS \cite{alnaes2015fenics}. All computations related to PCE and \PC{} were performed on a single NVIDIA A100 GPU on the Rockfish HPC facility at Johns Hopkins University. FEM simulations were carried out on a Standard Compute node at Rockfish, equipped with two Intel Xeon Gold 6248R “Cascade Lake” processors (24 cores per CPU, 3.0 GHz base clock) and 192 GB DDR4 RAM.

\begin{table}[ht]
\centering
\caption{Performance of PCE for operator learning and UQ across different example problems.}
\begin{tabular}{lcccccc}
\toprule
\multirow{3}{*}{Problem}
  & \multicolumn{2}{c}{Operator Learning}
  & \multicolumn{4}{c}{Uncertainty Quantification} \\
 \cmidrule(lr){2-3} \cmidrule(lr){4-7}
& MSE & Training
& \multicolumn{2}{c}{MAE}
& \multicolumn{2}{c}{Comp. time (s)} \\
\cmidrule(lr){4-5} \cmidrule(lr){6-7}
 & & time (s) & $\mu$ & $\sigma$ & MCS & \PC{} \\
\midrule
Anti-derivative        & 4.55e-9 & 0.3  & 1.50e-3 & 2.10e-3  & 2.02   & 1 \\
Advection-Diffusion    & 7.01e-8 & 0.4    & 1.58e-4 & 4.05e-5  & 1620   & 1.6   \\
1D Burgers          & 3.38e-7 & 0.5     & 8.88e-5 & 1.11e-4  & 1737   & 1.7   \\
2D Heat                & 6.21e-4 & 0.8     & 6.48e-16 & 6.19e-4  & 97200  & 56   \\
\bottomrule
\end{tabular}\label{tab:PCE-performance}
\end{table}

\begin{table}[ht]
\centering
\caption{Performance of \PC{} for operator learning and UQ across different example problems.}
\begin{tabular}{lcccccc}
\toprule
\multirow{3}{*}{Problem}
  & \multicolumn{2}{c}{Operator Learning}
  & \multicolumn{4}{c}{Uncertainty Quantification} \\
 \cmidrule(lr){2-3} \cmidrule(lr){4-7}
& MSE & Training
& \multicolumn{2}{c}{MAE}
& \multicolumn{2}{c}{Comp. time (s)} \\
\cmidrule(lr){4-5} \cmidrule(lr){6-7}
 & & time (s) & $\mu$ & $\sigma$ & MCS & \PC{} \\
\midrule
Anti-derivative       & 1.51e-8 & 0.5   & 2.21e-3 & 2.09e-3  & 2.02   & 1.6 \\
Advection-Diffusion    & 4.79e-5 & 30     & 4.68e-3 & 1.31e-3  & 1620   & 32   \\
1D Burgers            & 4.41e-5 & 76     & 2.08e-3 & 1.1e-3  & 1737   & 84   \\
2D Heat               & 6.24e-4 & 14     & 1.09e-6 & 5.3e-4  & 97200  & 116   \\
\bottomrule
\end{tabular}\label{tab:PC^2-performance}
\end{table}

\subsection{Anti-derivative example}

This example is adapted from the original DeepONet paper \cite{lu2021learning}. Here, we consider the problem of learning an anti-derivative operator. 
Let $u:[0,1] \rightarrow \mathbb{R}$ be an input function. The operator $\pazocal{G}$ maps $u$ to its antiderivative:

\begin{equation}
\begin{aligned}
    \dfrac{d}{dx} s(x) &= u(x), \quad x \in [0, 1], \quad \text{with } s(0) = 0, \\
    \Rightarrow \quad \pazocal{G}(u(x))(x) &= s(x) = \int_0^x u(t)\, dt,
\end{aligned}
\end{equation}
where  $u(x) \sim \operatorname{GP}\left(0, \operatorname{Cov}\left(x_1, x_2\right)\right)$. The covariance function is defined by the radial basis functions (RBF) kernel, $k\left(x, x^{\prime}\right)=\sigma^2 \exp \left(-\dfrac{\left(x-x^{\prime}\right)^2}{2 \ell^2}\right)$, with $\sigma=1$ and $\ell=0.2$. The random field is discretized using six KL terms ($r=6$), which capture $99\%$ of its total variance. We set the polynomial orders to $p=3$ and $q=10$ for both PCE and \PC{}. For PCE, $N=100$ input samples with corresponding output solutions are used as labeled training data, while $N=1000$ samples are reserved for testing. As shown in Table \ref{tab:PC^2-performance}, PCE provides excellent performance in terms of both accuracy and computational efficiency for both OL and UQ. 
\begin{figure}[!ht]
   \centering 
    \begin{tabular}{c}
\includegraphics[width=0.9\textwidth]{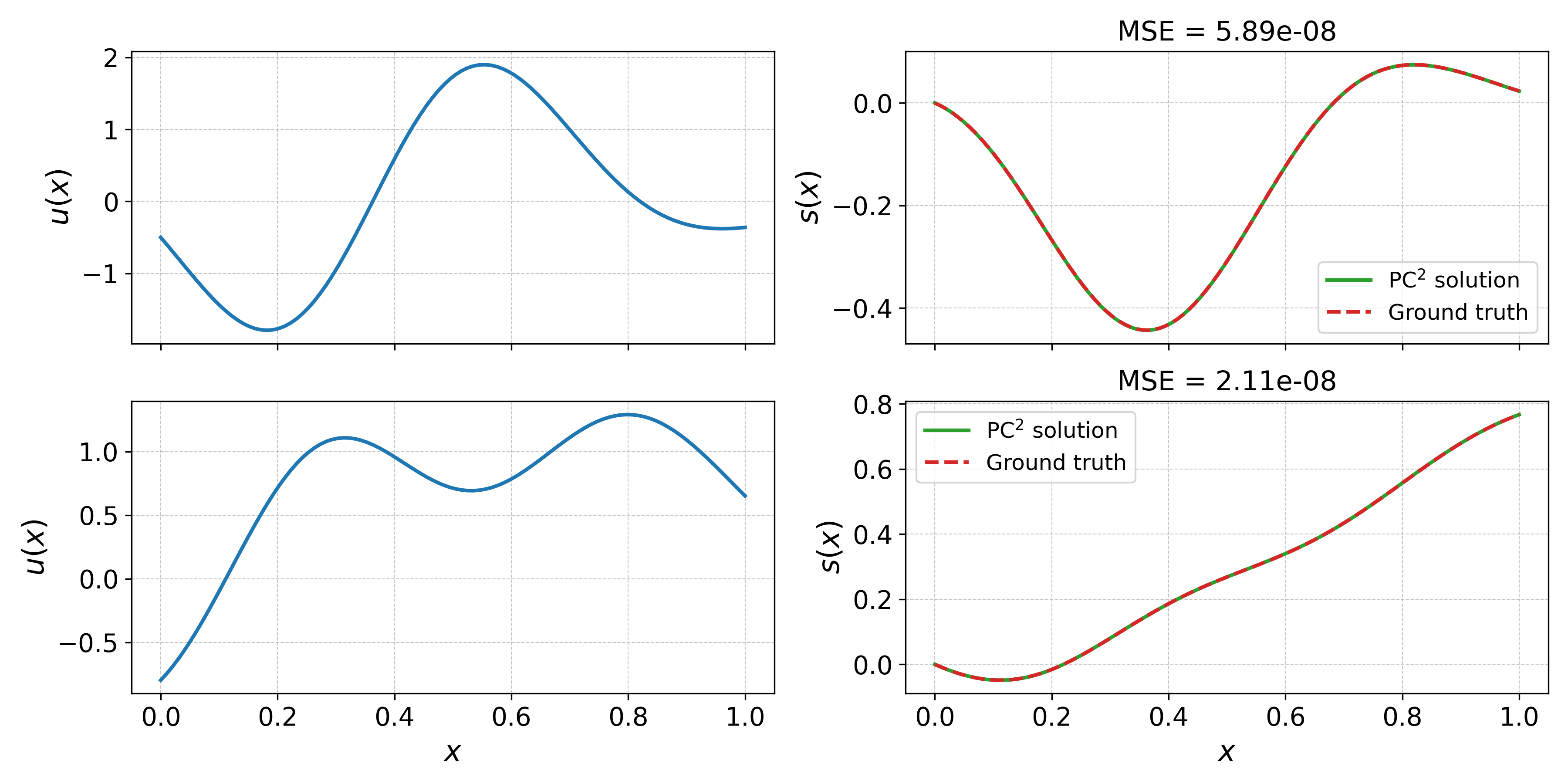}
\end{tabular}
\caption{OL plots for the anti-derivative example: Comparison of the \PC{} solution with the ground truth for different realizations of the input function $u(x)$. \textit{Top panel}: Worst-case realization with the highest prediction error. \textit{Bottom panel}: Sample at the 75\textsuperscript{th} percentile of the error distribution (i.e., $75\%$ of the cases exhibit lower error). The plots demonstrate the excellent performance of \PC{} in OL without requiring training data.}
\label{fig:anti-derivative OL}
\end{figure}

\begin{figure}[!ht]
   \centering 
    \begin{tabular}{c}
\includegraphics[width=0.95\textwidth]{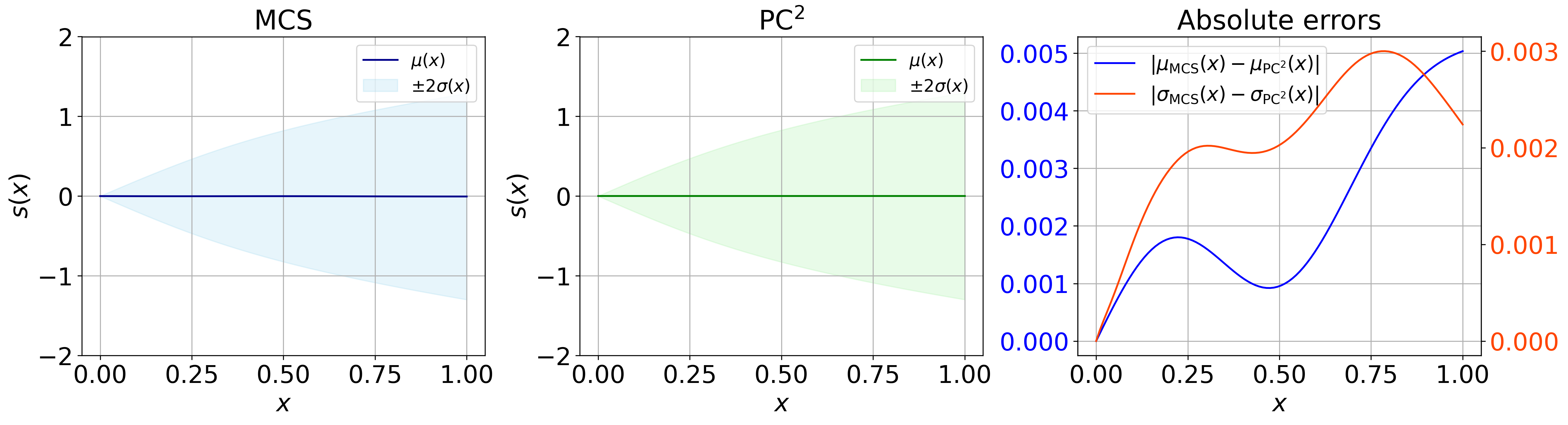}
\end{tabular}
\caption{UQ plots for the anti-derivative example: Comparison plots of the predicted mean $\mu(x)$ and the corresponding $\pm 2\sigma(x)$ confidence bounds obtained using both MCS and \PC{}, along with the absolute errors in mean and standard deviation. The plots highlight the strong UQ capabilities of \PC{}, achieved without requiring any labeled training data.}
\label{fig:anti-derivative UQ}
\end{figure}

For \PC{}, we generate 1,000 input samples $u(x)$ for training without access to labeled outputs, i.e., relying solely on the PDE constraints. The performance is tested on the same testing dataset as PCE. As previously noted, only the \PC{} results are used for visualization purposes. The OL plots are shown in Figure \ref{fig:anti-derivative OL}. For plotting, we select two representative input samples from the test set: one corresponding to the worst prediction error, and the other to the 75\textsuperscript{th} percentile error. As shown in Figure \ref{fig:anti-derivative OL}, the \PC{} provides nearly identical results to the ground truth, with MSE on the order of $10^{-8}$ for both cases. The training time for \PC{} is 0.5 s, highlighting its excellent computational performance in the OL task. Figure \ref{fig:anti-derivative UQ} shows the UQ plots for this problem, including the predicted mean along with the $\pm 2\sigma$ confidence bounds for both MCS and \PC{}, as well as the absolute errors in the mean and standard deviation. From the plots, it is evident that \PC{} provides excellent estimation of the mean and standard deviation as compared to MCS. The total computation time is 1.6 s, which includes preprocessing, training, and postprocessing. From this example, it is quite clear that \PC{} performs both the OL and UQ tasks with high accuracy and exceptional computational efficiency. 

\subsection{Advection-Diffusion equation}

In this example, we consider the 1D advection–diffusion equation with spatially varying advection velocity.
\begin{equation}
\begin{gathered}
\dfrac{\partial s(x, t)}{\partial t}+v(x) \dfrac{\partial s(x, t)}{\partial x}=D \dfrac{\partial^2 s(x, t)}{\partial x^2},\quad x \in[0,1], t \in[0,1] \\
\mathrm{IC}: s(x, 0)=\sin (\pi x) \\
\mathrm{BC}: s(0, t)=0,\ s(1, t)=0
\end{gathered}
\end{equation}
where $D$ is the diffusion coefficient and the parameter $v(x)$ is the advection velocity.

Here, we consider $D=0.1$ and sample $v(x)$ from a Gaussian random field (GRF) with mean velocity, $\mu = 1$, and covariance function defined using RBF kernel with $\sigma = 0.05$ and $\ell=0.2$. We use six KL terms ($r=6$) to discretize the random field, capturing $99\%$ of its total variance. The given equation is advection-dominated, with a mean Péclet number, $\mathrm{Pe} = \frac{\mu L}{D}$ = 10, which is often challenging to model using physics-informed machine learning approaches \cite{rout2021numerical}. We first learn the operator that maps the advection velocity $v(x)$ to the solution $u(x)$ using PCE and \PC{} and then perform UQ. We use polynomial orders as $p=3$ and $q=14$. Similar to the previous example, for PCE, we generate $N=100$ labeled training samples and use $N=1000$ samples for testing. As observed in Table \ref{tab:PCE-performance}, PCE achieves exceptional numerical accuracy with MSE on the order of $10^{-8}$ with just a fraction of a second of training time, highlighting its effectiveness in OL. We get similar performance in UQ, with MAE on the order of $10^{-4}$ for mean and $10^{-5}$ for standard deviation, while requiring only a fraction of the computational time compared to MCS.
\begin{figure}[!ht]
   \centering 
    \begin{tabular}{c}
\includegraphics[width=0.95\textwidth]{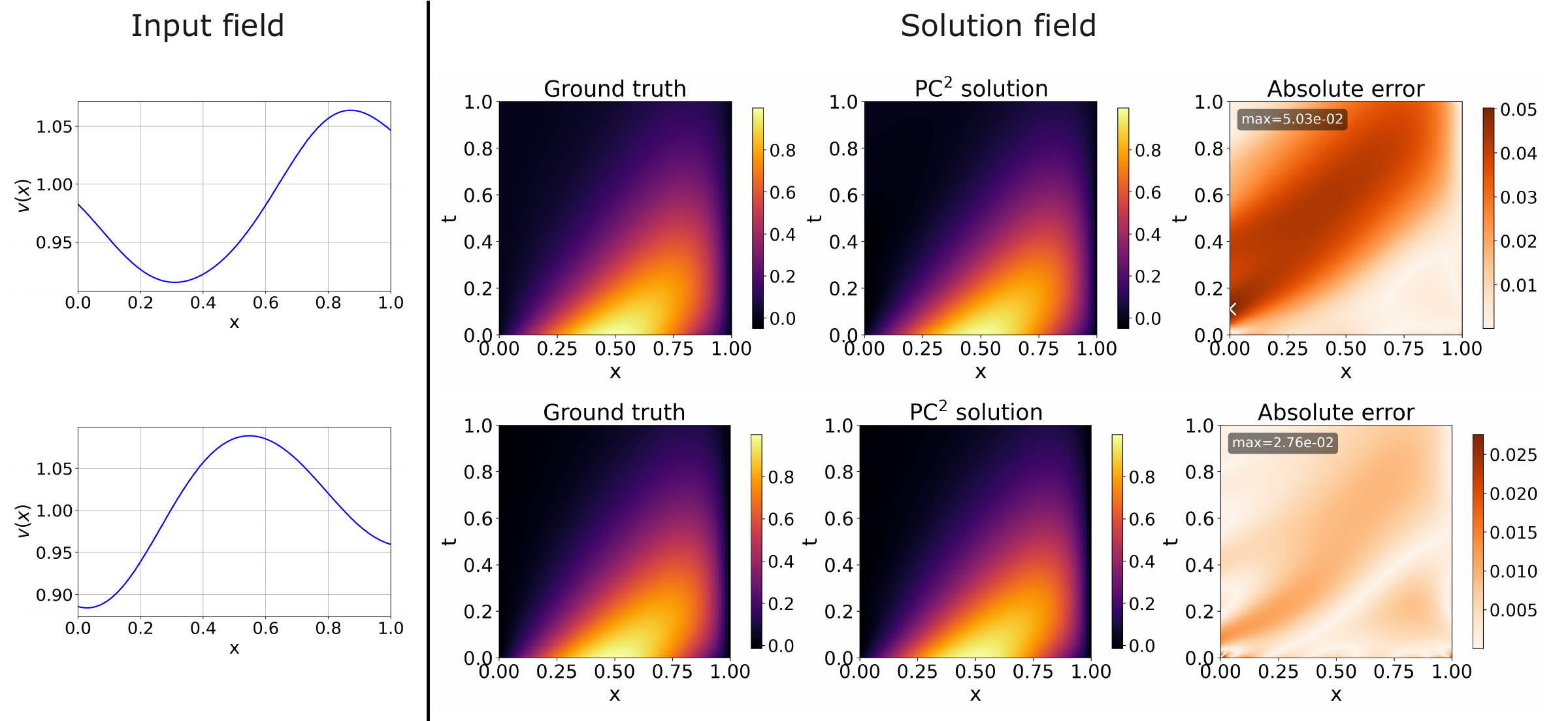}
\end{tabular}
\caption{OL plots for the 1D advection-diffusion equation: Comparison of the \PC{} solution with the ground truth, along with the corresponding absolute error for different realizations of the input field. \textit{Top panel}: Worst-case realization with the highest prediction error, corresponding to an MSE of $8.12\times 10^{-4}$. \textit{Bottom panel}: Sample at the 75\textsuperscript{th} percentile of the error distribution (i.e., $75\%$ of the cases exhibit lower error), with an MSE of $3.54 \times 10^{-5}$. The plots demonstrate the excellent performance of \PC{} in OL without requiring training data.}
\label{fig:advection-diffusion OL}
\end{figure}

For \PC{}, we generate 1,000 input samples of $v(x)$ for training without using the labeled outputs. We use the same testing dataset as PCE for evaluating its performance. Figure \ref{fig:advection-diffusion OL} shows the OL plots with test samples corresponding to the worst error, and the 75\textsuperscript{th} percentile error. As observed, the \PC{} solution closely matches the FEM solution even in the worst case scenario, with an MSE on the order of $10^{-4}$. For the 75\textsuperscript{th} percentile case, the MSE is on the order of $10^{-5}$, indicating excellent numerical accuracy in the operator learning (OL) task. In both cases, the maximum error occurs near the boundary. As shown in Table \ref{tab:PC^2-performance}, the training time for \PC{} in this example is 30 s, demonstrating superior computational efficiency.

\begin{figure}[!ht]
   \centering 
    \begin{tabular}{c}
\includegraphics[width=0.90\textwidth]{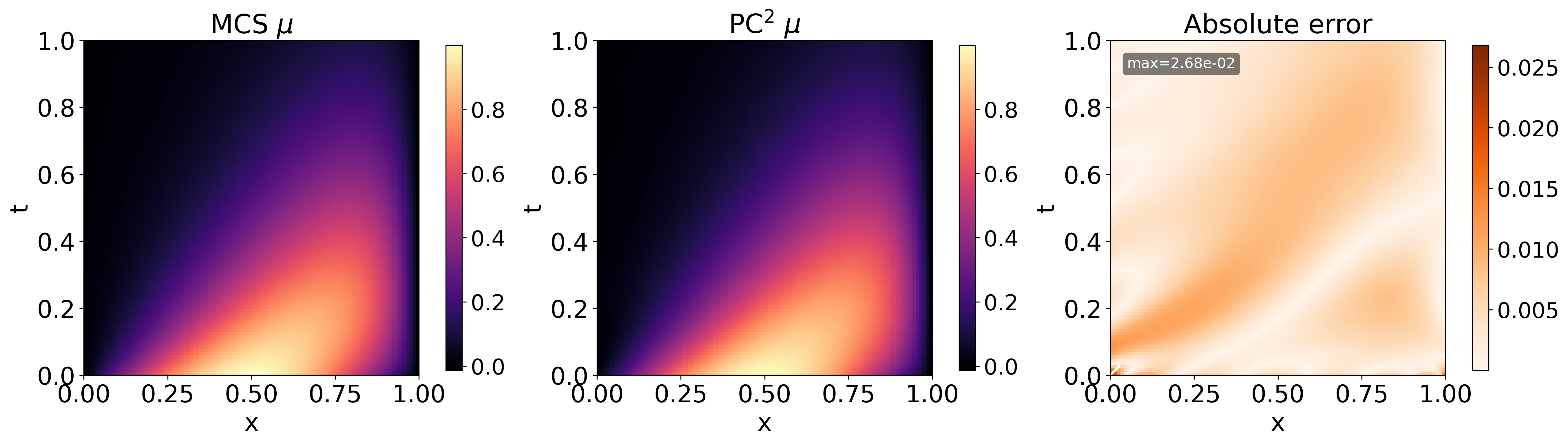}\\
\includegraphics[width=0.90\textwidth]{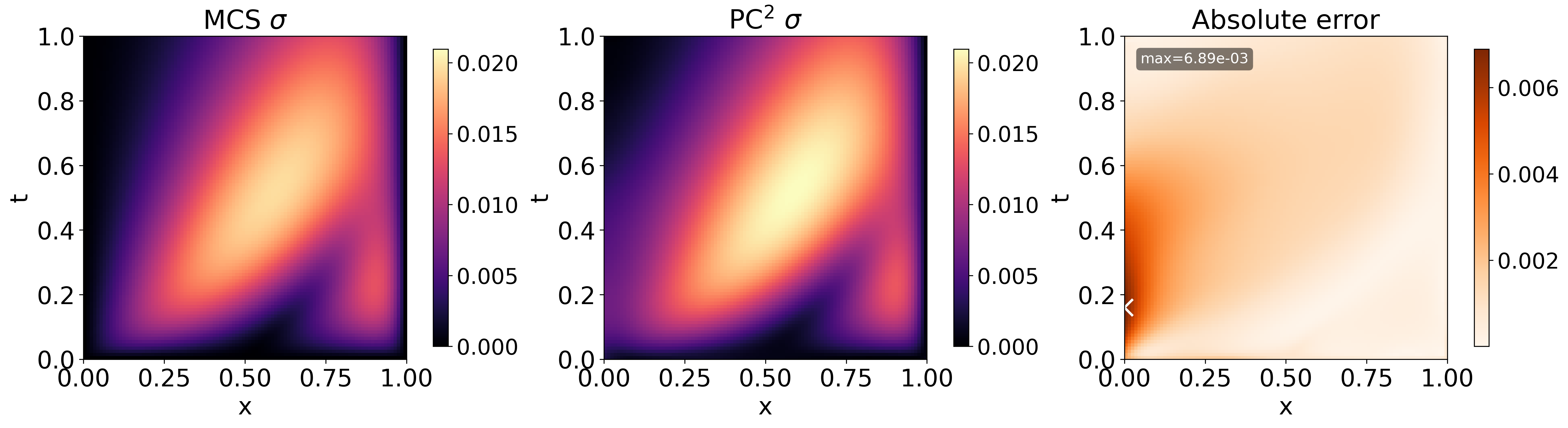}
\end{tabular}
\caption{UQ plots for the 1D advection-diffusion equation: \textit{Top panel}: Plots of the mean $\mu_s(x,\ t)$ obtained by MCS and \PC{}, along with their absolute error. \textit{Bottom panel}: Plots of the standard deviation of $\sigma_s(x,\ t)$ obtained by MCS and \PC{}, along with their absolute error. The plots highlight the strong UQ capabilities of \PC{}, without requiring training data.}
\label{fig:advection-diffusion UQ}
\end{figure}

Figure \ref{fig:advection-diffusion UQ} shows the UQ plots for this problem, including the mean and standard deviation over the test domain for both MCS and \PC{}, along with the corresponding pointwise absolute errors. It can be observed from Figure \ref{fig:advection-diffusion UQ} that \PC{} provides accurate estimates of mean and standard deviation across the entire domain, compared to MCS with low absolute errors. The maximum absolute errors in the mean and standard deviation occur at the boundary and remain relatively low, indicating high numerical accuracy over the given domain. The computation time for this problem is 32 s, compared to 1620 s for MCS, as shown in Table \ref{tab:PC^2-performance}, highlighting the superior computational efficiency of the proposed method. The results clearly demonstrate that \PC{} performs exceptionally well in both OL and UQ tasks.

\subsection{1D Burgers' equation example}

This example highlights the ability of \PC{} to learn nonlinear operators.
Consider the following 1D viscous Burgers’ equation with spatially varying source term as
\begin{equation}
\begin{gathered}
\dfrac{\partial s(x, t)}{\partial t}+s(x, t) \dfrac{\partial s(x, t)}{\partial x}- \nu \dfrac{\partial^2 s(x, t)}{\partial x^2}=f(x), \quad x \in[0,1], t \in[0,0.3] \\
\mathrm{IC}: s(x, 0)=\sin (\pi x) \\
\mathrm{BC}: s(0, t)=0,\ s(1, t)=0
\end{gathered}
\end{equation}
where $\nu$ is the fluid viscosity and $f(x)$ is the source term.

In this example, we set $\nu = 0.001$, and the source term $f(x)$ is drawn from a zero-mean GRF with an RBF kernel, with $\sigma=0.1$ and $\ell=0.2$. Six KL terms ($r=6$) are used to discretize the random field, accounting for $99\%$ of its variance. We set polynomial orders as $p=3$ and $q=23$. Following the approach in the previous example, we generate $N=100$ labeled training samples and $N=1000$ samples for testing. As shown in Table \ref{tab:PCE-performance}, PCE demonstrates exceptional performance in OL, achieving the MSE on the order of $10^{-7}$ with a training time of just a fraction of a second. Similarly, we get excellent performance in UQ, with MAE on the order of $10^{-5}$ for the mean and $10^{-4}$  for the standard deviation, at a fraction of the computational cost compared to MCS.
\begin{figure}[!ht]
   \centering 
    \begin{tabular}{c}
\includegraphics[width=0.95\textwidth]{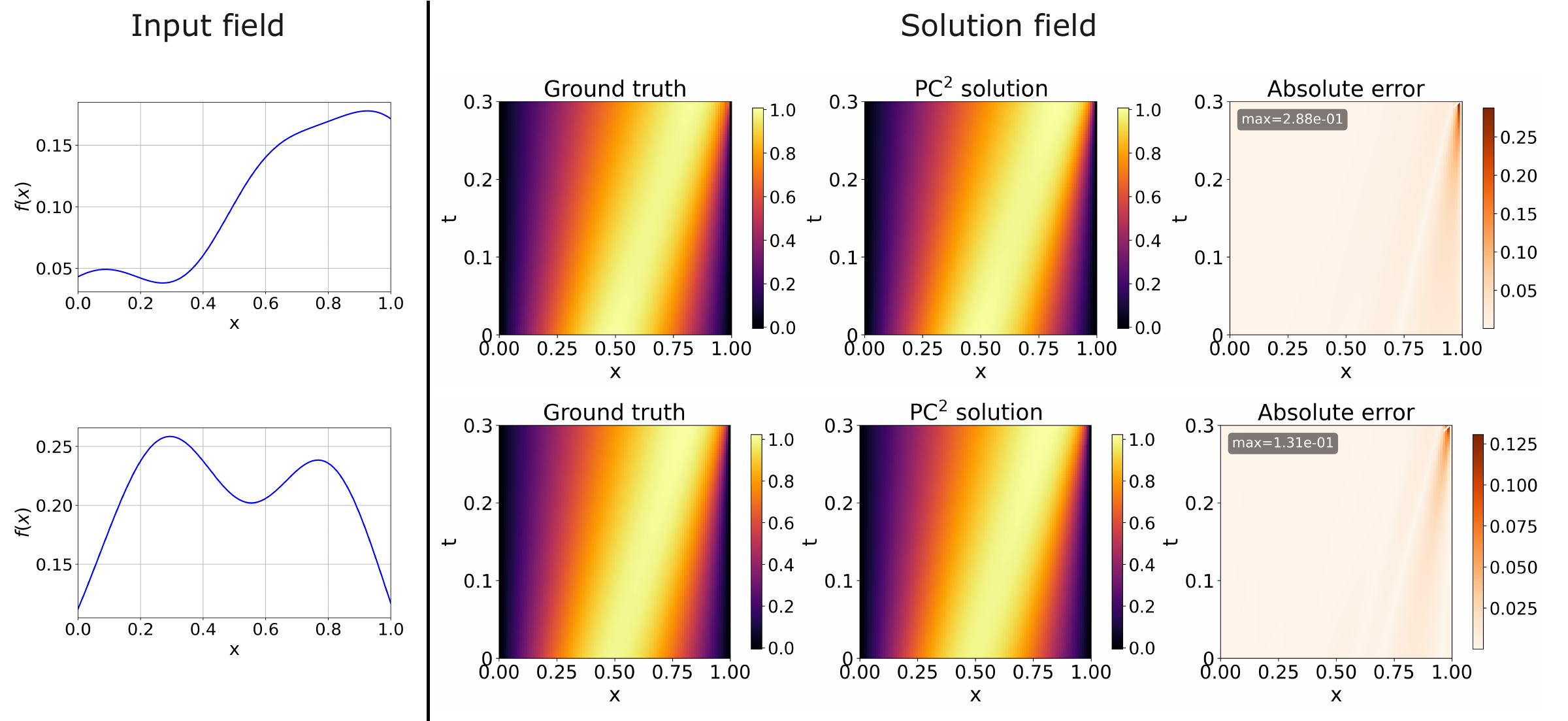}
\end{tabular}
\caption{OL plots for the 1D Burgers' equation: Comparison of the \PC{} solution with the ground truth, along with the corresponding absolute error for different realizations of the input field. \textit{Top panel}: Worst-case realization with the highest prediction error, corresponding to an MSE of $2.82\times 10^{-4}$. \textit{Bottom panel}: Sample at the 75\textsuperscript{th} percentile of the error distribution (i.e., $75\%$ of the cases exhibit lower error), with an MSE of $4.15 \times 10^{-5}$. The plots demonstrate the excellent performance of \PC{} in learning nonlinear operators.}
\label{fig:burgers OL}
\end{figure}

For \PC{}, we use 1,000 samples of $f(x)$ for training without using the labeled outputs, and the performance is tested on the same testing dataset as PCE. Figure \ref{fig:burgers OL} shows the OL results corresponding to the worst case and the 75\textsuperscript{th} percentile case. As shown in Figure \ref{fig:burgers OL}, \PC{} yields highly accurate results, with a low worst-case MSE of $2.82 \times 10^{-4}$. This highlights the strong capability of the proposed method in learning a nonlinear operator. Furthermore, the maximum error in both cases is observed near the right boundary at $t=0.3$, corresponding to the steepening phase of the solution where sharp gradients emerge.
\begin{figure}[!ht]
   \centering 
    \begin{tabular}{c}
\includegraphics[width=0.90\textwidth]{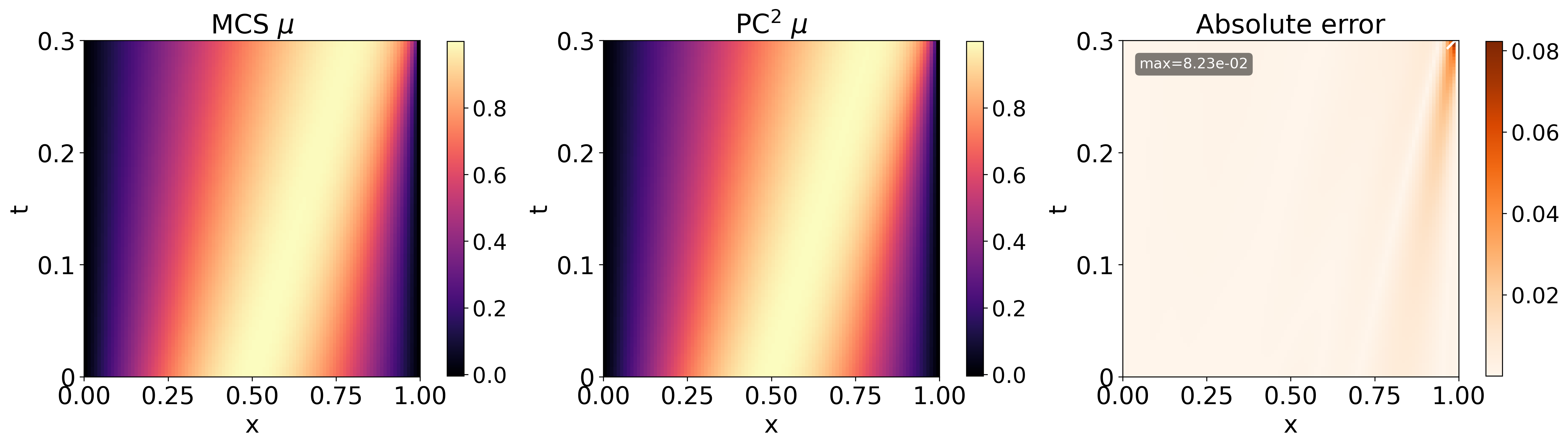}\\
\includegraphics[width=0.90\textwidth]{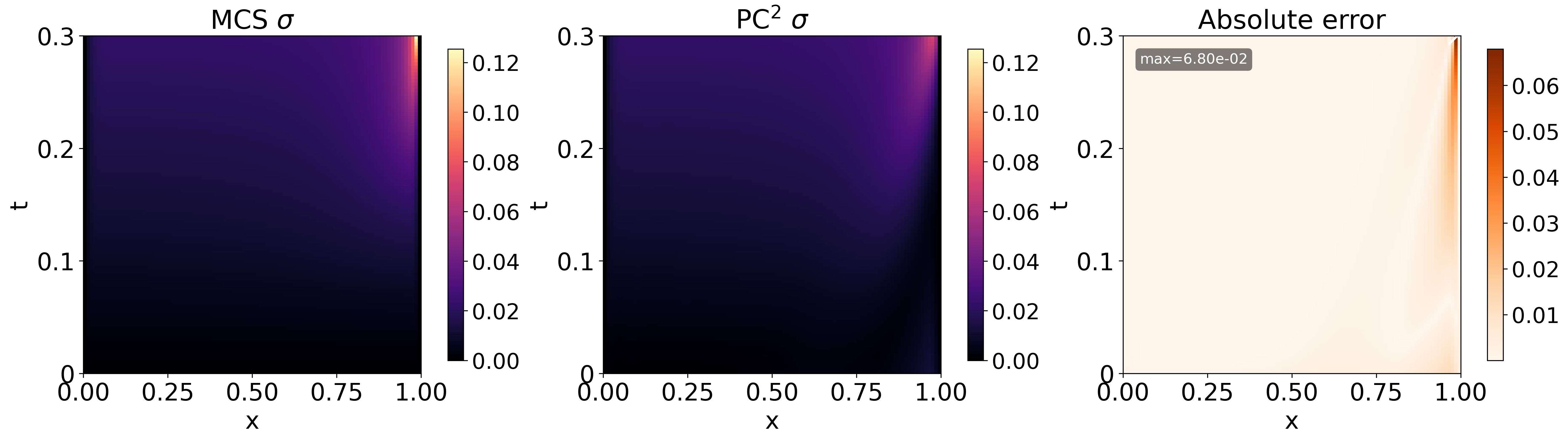}
\end{tabular}
\caption{UQ plots for the 1D Burgers' equation: \textit{Top panel}: Plots of the mean $\mu_s(x,\ t)$ obtained by MCS and \PC{}, along with their absolute error. \textit{Bottom panel}: Plots of the standard deviation of $\sigma_s(x,\ t)$ obtained by MCS and \PC{}, along with their absolute error. The plots highlight the strong UQ capabilities of \PC{}.}
\label{fig:burgers UQ}
\end{figure}

Figure \ref{fig:burgers UQ} shows the UQ plots, where the estimated mean and standard deviation obtained using \PC{} are compared with their reference counterparts from MCS. It can be observed from Figure \ref{fig:burgers UQ} that \PC{} provides accurate estimates of the mean and standard deviation over the given domain, as evidenced by the absolute error plots, with a maximum error on the order of $10^{-2}$. To achieve this level of accuracy, \PC{} requires only 80 s of computation time, compared to 1737 s for MCS, as shown in Table \ref{tab:PC^2-performance}, highlighting its superior computational efficiency. These results demonstrate the superior performance of PCE in UQ.

\subsection{2D Heat equation}

Consider the following 2D Heat equation with a spatially varying source term and initial condition as
\begin{equation}
\begin{array}{cc}
\dfrac{\partial s(x, y, t)}{\partial t}-\alpha\left(\dfrac{\partial^2 s(x, y, t)}{\partial x^2}+\dfrac{\partial^2 s(x, y, t)}{\partial y^2}\right) = f(x, y), & (x, y) \in(0,1) \times(0,1), t \in[0,1], \\

\mathrm{IC}: s(x, y, 0)=A \sin (2 \pi x) \sin (2 \pi y), & (x, y) \in[0,1] \times[0,1], \\

\mathrm{BC}: s(x, y, t)=0, & (x, y) \in \partial([0,1] \times[0,1]) .
\end{array}
\end{equation}
where $\alpha$ is the thermal diffusivity of the medium and $f(x,y)$ is the source term. 
\begin{figure}[!ht]
   \centering 
    \begin{tabular}{c}
\includegraphics[width=0.95\textwidth]{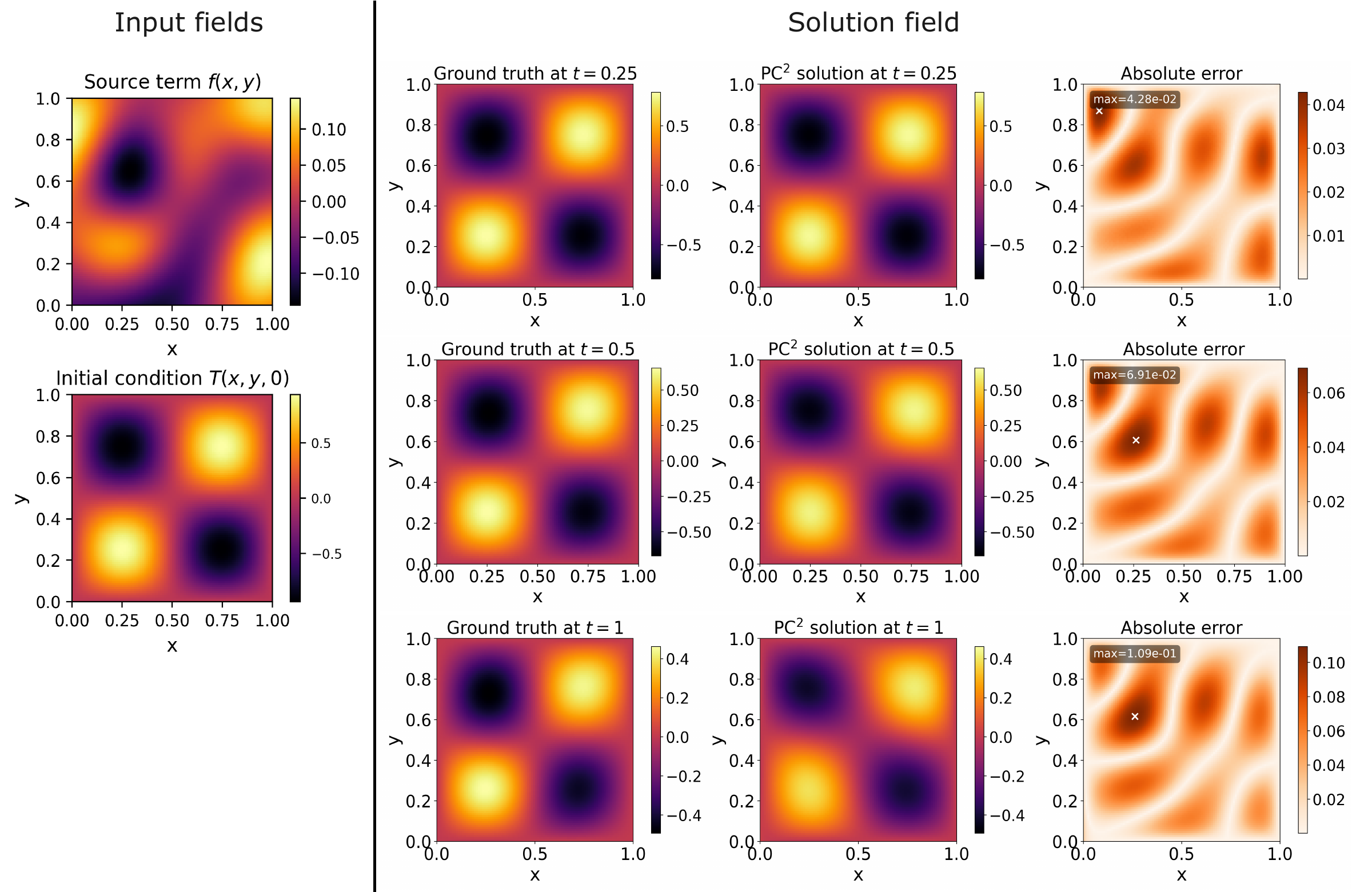}
\end{tabular}
\caption{OL plots for the 2D heat equation: Comparison of the \PC{} solution with the ground truth, along with the corresponding absolute error at different time stamps for the 75\textsuperscript{th} percentile case. The MSE for this case is $7.83 \times 10^{-4}$. The plots demonstrate the excellent performance of \PC{} in OL for high-dimensional problems.}
\label{fig:2D heat OL}
\end{figure}

Here, we consider $\alpha=0.01$ and the source term $f(x,y)$ is sampled from a zero-mean 2D GRF with an RBF covariance kernel, using $\sigma = 1$ and $\ell=0.2$. The random field is discretized using a Karhunen–Loève (KL) expansion with 20 KL terms, capturing $99\%$ of the total variance. The amplitude ($A$) of the initial condition is modeled as a standard normal random variable. So, in total, we have 21 stochastic variables ($r=21$). We use a polynomial order of $p = 4$ for the stochastic basis with a hyperbolic truncation factor of 0.9 \cite{blatman2011adaptive}, and set the spatial-temporal expansion order to $q = 16$. For training, we use $N = 2500$ samples, and for testing, $N = 100$ samples are employed. As observed in Table \ref{tab:PCE-performance}, PCE provides high numerical accuracy and computational efficiency for OL. For UQ, we compute the MAE in mean and standard deviation with respect to MCS at $t=1$. The MAE is on the order of $10^{-16}$ for the mean and $10^{-4}$ for the standard deviation. The computation time is just 56 s compared to $\sim$27 h for MCS, demonstrating exceptional computational efficiency. 

For \PC{}, we use 5,000 input samples without labeled solutions for training, and test the performance on the testing dataset as PCE. Figure \ref{fig:2D heat OL} presents the OL results for a test sample corresponding to the 75\textsuperscript{th} percentile error. We plotted the results for different time stamps, i.e., $t=0.25, 0.5,\text{and}\ 1$. As observed, the MSE for this case is $7.83 \times 10^{-4}$, which suggests high numerical accuracy. The error tends to increase with time, as evident from the absolute error plots with a maximum error of $1.09 \times 10^{-1}$ for $t=1$. As given in Table \ref{tab:PC^2-performance}, the total MSE is  $6.2 \times 10^{-4}$, and \PC{} achieves this error with only 14 s of training time, which is exceptional considering high stochastic dimensions ($r=21$) with three physical variables. This highlights the scalability of the proposed method. Furthermore, \PC{} achieves accuracy identical to standard PCE, but without the need for training data, which is computationally expensive to generate for this example. This highlights the superior performance of \PC{}, particularly for high-dimensional problems where acquiring labeled training data is computationally expensive.

\begin{figure}[!ht]
   \centering 
    \begin{tabular}{c}
\includegraphics[width=0.90\textwidth]{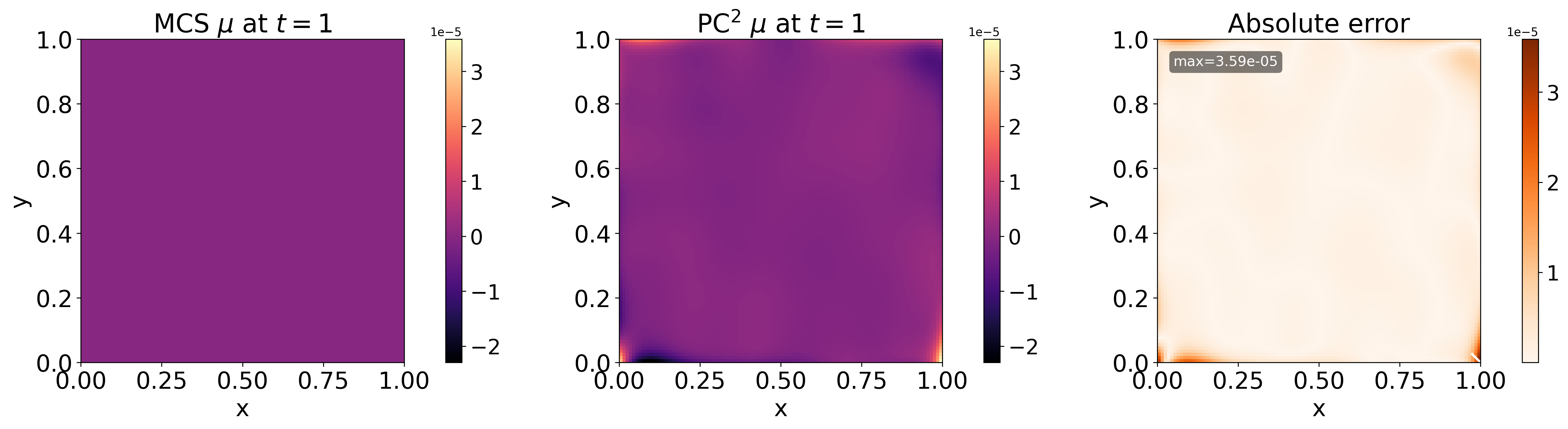}\\
\includegraphics[width=0.90\textwidth]{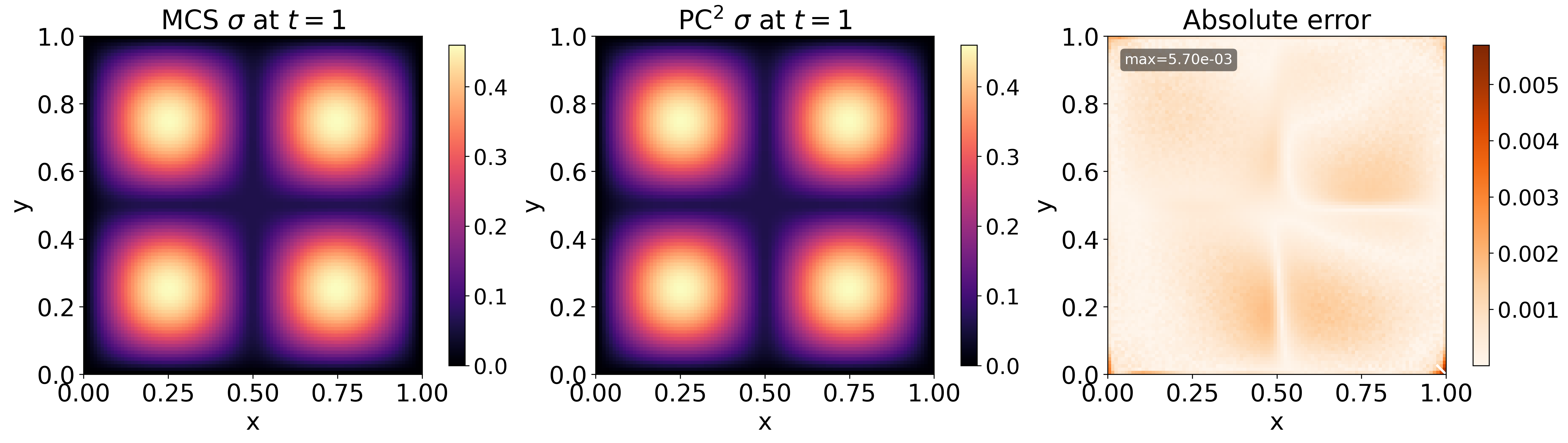}
\end{tabular}
\caption{UQ plots for the 2D heat equation: \textit{Top panel}: Plots of the mean $\mu_s(x,\ t)$ obtained by MCS and \PC{} at $t=1$ , along with their absolute error. \textit{Bottom panel}: Plots of the standard deviation of $\sigma_s(x,\ t)$ obtained by MCS and \PC{} at $t=1$, along with their absolute error. The plots highlight the strong UQ capabilities of \PC{}, without requiring training data.}
\label{fig:2D heat UQ}
\end{figure}

Figure \ref{fig:2D heat UQ} shows the UQ plots for the given problem. We obtain the estimated mean and standard deviation at $t=1$ and compare them with their reference counterparts from MCS. It can be seen from the plots that \PC{} provides excellent estimates of the mean and standard deviation, with maximum absolute errors of $3.59 \times 10^{-5}$ for the mean and $5.7 \times 10^{-3}$ for the standard deviation. The computation time for \PC{} is just 116 s, compared to $\sim$27 h for MCS, as shown in Table \ref{tab:PC^2-performance}, highlighting the significant computational efficiency..

\section{Conclusion}

In this work, we have proposed a polynomial chaos expansion (PCE) framework for operator learning in a data-driven and physics-informed setting. In the proposed framework, the task of operator learning reduces to solving a standard system of linear/nonlinear equations. Furthermore, the framework provides uncertainty quantification at no extra computational cost. We have demonstrated the effectiveness of the proposed method in OL and UQ-related tasks by solving linear/nonlinear PDE problems with various input functions, including source terms, parameters, and initial conditions. Numerical results indicate that the proposed method provides high accuracy at minimal computational cost and resource requirements, highlighting its efficiency, practicality, and scalability. 


\section{Acknowledgements}
This work was supported by the Defense Threat Reduction Agency, Award HDTRA12020001. LN gratefully acknowledges the support of the Czech Science Foundation under project No. 24-11845S. The international collaboration was partly supported by the Ministry of Education, Youth and Sports of the Czech Republic under project No. LUAUS24260.

\bibliographystyle{ieeetr}
\bibliography{references}  
\end{document}